\documentclass[conference]{IEEEtran}
\usepackage[dvips]{graphicx}

\usepackage{cite}

\ifCLASSINFOpdf
\else
\fi
\usepackage[cmex10]{amsmath}
\usepackage{algorithmic}

\usepackage{array}

\RequirePackage{algorithm}
\usepackage{amssymb}

\usepackage{fancyhdr}
\begin{document}
%
\title{A Tunable Particle Swarm Size Optimization Algorithm for Feature Selection}

\author{\IEEEauthorblockN{Naresh Mallenahalli$^1$, T. Hitendra Sarma $^2$}
\IEEEauthorblockA{$^1$Scientist, Software and Database Systems Group, \\ National Remote Sensing Center (ISRO), Hyderabad, Andhra Pradesh, 500 625, India. \\ e-mail: nareshkumar\_m@nrsc.gov.in\\ 
$^2$ Srinivasa Ramanujan Institute of Technology,\\ Anantapur, Andhra Pradesh, 515 701, India. \\ e-mail:t.hitendrasarma@gmail.com }}



%


\maketitle

\begin{abstract}
Feature selection is the process of identifying statistically most relevant features to improve the predictive capabilities of the classifiers. To find the best features subsets, the population based approaches like Particle Swarm Optimization(PSO) and genetic algorithms are being widely employed. However, it is a general observation that not having right set of particles in the swarm may result in sub-optimal solutions, affecting the accuracies of classifiers. To address this issue, we propose a novel tunable swarm size approach to reconfigure the particles in a standard PSO, based on the data sets, in real time. The proposed algorithm is named as Tunable Particle Swarm Size Optimization Algorithm (TPSO). It is a wrapper based approach wherein an Alternating Decision Tree (ADT) classifier is used for identifying influential feature subset, which is further evaluated by a new objective function which integrates the Classification Accuracy (CA) with a modified F-Score, to ensure better classification accuracy over varying population sizes. Experimental studies on bench mark data sets and Wilcoxon statistical test have proved the fact that the proposed algorithm (TPSO) is efficient in identifying optimal feature subsets that improve classification accuracies of base classifiers in comparison to its standalone form.
\footnote{This paper is accepted for oral presentation at IEEE Congress on Evolutionary Computation (CEC) - WCCI 2018, Rio de Janerio, Brazil,  \#18120}

\end{abstract}
 
\vspace{0.2in}
\begin{keywords}
Feature Selection, Evolutionary Computing, Particle Swarm Optimization, Alternate Decision Trees.
\end{keywords}



%
\IEEEpeerreviewmaketitle

\section{Introduction}

In statistical pattern recognition, each pattern represents a real world object described by a set of features (synonymously called as dimensions, here after). More the number of features used, better the description of the object. However, all the features may not be important for the decision making problem on hand. For instance, a {\bf student} can be described with the features like {\em height, weight, regularity, father name, family income, etc}. Now, for the problem on hand, like selecting a student for a basket ball team, the feature like {\em height} and {\em weight} are highly {\bf relevant}, where as the features like {\em father name}, {\em family income} are {\bf irrelevant}. The features, {\em regularity} and {\em family income} are highly {\bf relevant} to classify whether the student shall be awarded the fellowship or not. Hence, best feature selection for the problem on hand is important for quality decision making.  

Now-a-days, with the development of high-throughput technologies, it is possible to measure hundreds of feature values for each object, which has resulted in large volumes of high dimensional data for analysis. In hyper spectral image analysis, using advanced hyper spectral instruments, hundreds of feature values (each one corresponding to one spectral band) can be measured for each object on earth ~\cite{bioucas2013hyperspectral}. In contemporary scientific applications, it is quite often to get such large volumes of high dimensional data, which becomes very challenging problem for analysis ~\cite{hua2009performance}. 

In pattern classification, irrelevant (some times redundant or noisy) features will affect the classification accuracy. It has been proved that, in the presence of large number of features, the learning models become overfit on the training data, which leads to poor generalizability of the trained model, offering a great challenge for pattern classification and prediction problems. Thus, the feature selection process has been considered as a pre-processing step to eliminate irrelevant and redundant features, which is critical for decision making in real world applications~\cite{liu2002,Guyon2003,Guyon2002,Liu2005,yang2012particle}.

The feature selection algorithms have been widely used in many application areas such as genomic analysis~\cite{Inza2004}, text classification~\cite{Forman2003}, information retrieval~\cite{Swets1995}, intrusion detection~\cite{Lee2000}, bio informatics~\cite{Saeys2007} etc. A comprehensive survey on feature selection methods is published in~\cite{chandrashekar2014survey}. Empirical studies on feature selection algorithms for real world problems are presented in~\cite{Tao2004,sun2005,Shuangge2006,Murie2009}. 

Feature selection is an optimization problem which aims to determine an optimal subset of $d$ features out of $n$ features in the input data $(d<<n)$, that maximize the classification or prediction accuracy. Performing an exhaustive search to find an optimal subset of $d$ features out of all possible $2^n$ candidate feature subsets, based on some {\em evaluation criterion}, is computationally infeasible, and it becomes an NP-hard problem with the increasing $n$ value~\cite{chandrashekar2014survey}.
Hence, different other search strategies like {\em complete}, {\em sequential}, {\em random search} are explored.
However, most of these approaches suffers from local minima problem. Therefore, Evolutionary Computation (EC) techniques, which ensures global optimum or near global optimum, such as Genetic Algorithms(GAs)~\cite{raymer2000dimensionality}, Genetic Programming (GP) and Particle Swarm Optimization (PSO), were used in many feature selection problems. As stated in ~\cite{xue2012new}, PSO is simple to understand and easy to implement than GP and GAs and able to handle optimization problems with multiple local optima reasonably well, it requires less number of parameters and can converge more quickly. However, the efficiency of PSO depends on various input parameters that are to be tuned properly ~\cite{shi1998},~\cite{angeline1998}. More detailed study on PSO and its improvements is presented in ~\cite{xue2016survey}.

In the standard PSO, the swarm size is an important parameter, where very small swarm size will lead to local minima, while large swarm size would slow down the algorithm~\cite{vandenBergh2001}. To address this issue, in the present work, we intend to vary the population sizes of the particles in standard PSO based on the data sets in real time. A new objective function has been developed which integrates the accuracy of the classifier with the modified F-Score. Finally, we propose a new PSO search method for feature selection using {\em tunable swarm size configuration}. The efficiency of the proposed method is compared with other popular contemporary feature selection methods. 

This paper is organized as follows. Section ~\ref{Survey} presents the brief review of the existing methods for feature selection. 
In Section~\ref{pso1} we briefly outlined the standard PSO methodology and presented the motivation for the {\em tunable swarm size configuration} in the present work. Section~\ref{adt} outlines the Alternating Decision Tree classifier, which is used along with the standard PSO for feature subset selection. The proposed Tunable Particle Swarm Size Optimization (TPSO) algorithm is presented in Section~\ref{var-swarm}. 
The experiments and results are presented in Section~\ref{expresults}. 
Conclusions and discussion are presented in Section~\ref{concl}.

\section{Feature Selection Methods}\label{Survey}

In literature, the feature selection methods are broadly classified into three categories {\em viz.,} filter, wrapper and embedded methods. Filter methods select the feature based on the given data, irrespective of the classifier. In the wrapper model, feature selection will be done based on the feedback of the predefined learning model. Wrapper based methods will find better and optimal feature subsets with high accuracy, as they are considering the feedback of the learning model, but it requires expensive computation. 
However, it is proved that filters have better generalization capabilities than wrapper based ones~\cite{Inza2004}. 

Algorithms with embedded models such as C4.5\cite{Quinlan1993} and least angle regression (LARS)~\cite{Efron2004}, the variable or feature selection process is incorporated as part of the training process, and the relevance of the selected feature is analyzed by the objective function of the learning model under consideration. Both filter and embedded approaches may result a subset of selected features or the weights that represent the relevance or importance of all features. 

Some feature selection methods compute the ranks of all features using some ranking criterion, such methods are simple and computationally efficient.  These rank based methods are more robust against over fitting, resulting more bias with less variance~\cite{Guyon2003,Hastie2003}. Further, the statistical approaches such as T-Statistics, F- Statistics, Chi-square test etc., have been explored significantly in the literature~\cite{jin2006machine},~\cite{Wang2007}. A few other feature selection approaches are based on the concepts of information theory such as information gain~\cite{Liu2004}, mutual information~\cite{Guyon2003,peng2005}, and entropy-based measure~\cite{Liu2005}. 
Machine learning techniques including evolutionary algorithms, SVMs, Decision Trees etc., are also been used for feature selection.~\cite{Tan2008,winkler2012analysis,Malin2008}. More recently, the evolutionary Computing techniques such as such as Genetic Algorithms(GAs) and Particle Swarm Optimization (PSO) are being used popularly used for feature selection. Bing Xue {\em et.al.,} explored the performance of PSO and various other improvements in ~\cite{xue2016survey}. PSO is widely used for Feature Selection on High-dimensional Datasets~\cite{tran2016investigation}. A good survey on novel population topologies for improving the performances of population-based optimization algorithms for solving single objective optimization, multiobjective optimization and other classes of optimization problems is presented in~\cite{lynn2017population}.

This paper presents an improvement over the standard PSO, which is a wrapper based approach to improve the classification accuracy with reduced number of features.

\section{Standard Particle Swarm Optimization}\label{pso1}

Particle Swarm Optimization imitates the movement of a flock of birds, where each bird has its own intelligence to find the best direction to move and to reach the destination as a whole. In standard PSO, each single candidate solution is considered as a particle in the search space. 
For each particle, there is a fitness value, computed using a {\em fitness function} to be optimized, and velocity, which determine the movement of the particles. During movement, each particle updates its position based on its previous position, velocity and as well as considering the positions of neighbouring particle. 

The standard PSO starts with a randomly initialized population (particles) of size $N$. Each particle $P_{i}$ is identified as a point in the $d$ dimensional space $X_{i}=(x_{i1},x_{i2},\ldots,x_{id})$. $pbest$ represents the {\em fitness values} of the best positions of the particles given by $F_{i}=(f_{i1},f_{i2},\ldots,f_{id})$. $gbest$ represents the index of the particle that has the best fitness value in the swarm. The velocity of a particle $i$ is represented by $V_{i}=(v_{i1},v_{i2},\ldots,v_{id})$. 

The iterative approach starts with an initial random solutions (particles in initial swarm). 
In each iteration, for each particle, the velocity and the position are updated using the following equations:


 \begin{equation}
 x^{t}_{ij}=x^{t-1}_{ij}+v^{t}_{ij}
  \end{equation}
\begin{equation}
  v^{t}_{ij}=w*v^{t-1}_{ij}+\eta_{1}*r1()*(p_{ij}-x^{t}_{ij})+\eta_{2}*r2()*(p_{gj}-x_{ij}),
  \end{equation}

where $j=1,\ldots,d$, $w$ is a positive linear function of time which updates according to the generation iteration. The $\eta_{1}$ and $\eta_{2}$ represent the acceleration terms that pull the particles towards $pbest$ and $gbest$. 
The $r1()$ and $r2()$ are random number generation functions, which generates random values that are uniformly distributed in $[0,1]$.
The terms $p_{ij}$ and $p_{gj}$ represents the $pbest$ and $gbest$ in the $j^{th}$ dimension respectively.
The velocities of the particles are bounded by a maximum limit $V_{max}$. 
If $V_{max}$ is too small then it may end up with a local optima, and if the $V_{max}$ is too large then the particles may fly beyond the good solutions.

The swarm size is a critical parameter in this standard PSO algorithm wherein very few particles will make the algorithm to get stuck at the local optima, while too many particles would slow down the algorithm~\cite{vandenBergh2001}. It is the key factor that has motivated the present research work. 

In this paper, we propose a new particle swarm optimization search for feature subset selection using {\em tunable swarm size configuration}, which is explained in Section~\ref{var-swarm}. 

\section{Alternating Decision Trees (ADT)}\label{adt}

Alternating Decision Trees (ADT) are often considered as generalization of conventional decision trees~\cite{freund1999alternating}. ADT generates the decision rules based on majority voting taking all simple rules into account. It consists of decision nodes and prediction nodes. The prediction nodes contain a numeric value having a {\em positive} or {\em negative} sign, and the decision node specify a condition.  The decision nodes will be splitting nodes whereas the prediction nodes are either root or leaf nodes. An instance is classified by traversing from the root by following all paths where all the decision nodes are true. A {\em positive} sum of all prediction nodes that are been traversed implies the membership of one class and the {\em negative} sum implies the membership of other class. Empirical studies proved that, under some favourable conditions ADTs are more robust than the conventional decision trees, C4.5 and J48~\cite{naresh2013}.  

\section{Tunable Particle Swarm Size Optimization Algorithm (TPSO)} \label{var-swarm}

In this section we present our new algorithm called {\em Tunable Particle Swarm Size Optimization Algorithm} (TPSO) which will find the best initial swarm size for the given data to overcome the local minima problem~\cite{vandenBergh2001}.   
 
The data set is split into testing and training folds using a stratified $k$ fold cross validation procedure. For each of the training data sets we first initialize swarm size and then select the features using the standard PSO and Alternating Decision Tree (ADT). 
We then compute the test accuracy using the features subset identified in the previous step and ADT classifier. A new feature score which measures the discrimination between features having two sets of numbers categorical or numeric with respect to the decision attribute is then computed following the procedure in Section~\ref{newFDS}.

\subsection{New Feature Discrimination Score}
\label{newFDS}

Consider a given the data set $X_{ji}$, for $j=1,2,\ldots,n-1,n$, $i=1,2,\ldots,m$ having $n$ rows and $m$ features where the last feature is the decision class. If the decision class is binary then $n^{+}$ and $n^{-}$ denote positive and negative instances respectively. In~\cite{liu2011improved} a feature discriminatory score using mean of the attribute values is computed. In our approach instead of mean we employ median as it is the best representative the central tendency of data sets with skewed distribution. We define the feature score of the $i^{th}$ feature as:

\begin{equation}
\label{chap4:eqn1}
F(i)=\frac{V_1}{V_2}
\end{equation}

\begin{equation}
\label{chap4:eqn1-1}
V_1={abs((\tilde{x_{i}}^{+}-\tilde{x_{i}}{^2})+(\tilde{x_{i}}^{-}-\tilde{x_{i}}{^2}))}
\end{equation}

\begin{equation}
\label{chap4:eqn1-2}
V_2={\frac{1}{n^{+}-1}\sum_{k=1}^{n^{+}}(x_{k,i}^{+}-\tilde{x_{i}}^{+})^{2}+\frac{1}{n^{-}-1}\sum_{k=1}^{n^{-}}(x_{k,i}^{-}-\tilde{x_{i}}^{-})^{2}}
\end{equation}

where $\tilde{x_{i}}^{+}$ denotes the median of the values in the $i^{th}$ attribute corresponding to the positive decision class, $\tilde{x_{i}}^{-}$ denotes the median of the attribute values corresponding to the negative decision class, $\tilde{x_{i}}$ denotes the median of all the values of the $i^{th}$  attribute, $x_{k,i}^{+}$ is the median value of the $i^{th}$ feature of the $k^{th}$ positive instance and $x_{k,i}^{-}$ is the median value of the $i^{th}$ feature of the $k^{th}$ negative instance.

\subsection{Fitness function}
We develop a new fitness function to evaluate the effectiveness of the feature subsets as mentioned below.

\begin{equation}
V_{i}=0.5*A+0.5*(M_{1}/M_{2}) 
\end{equation}

where $A$ is the accuracy obtained using ADT, $M_{1}$ is the sum of the discriminatory scores of the features in the reduced feature subset, that is $M_1=\sum_{i=1}^{r} F(i)$ where $r<m$, $M_{2}$ is the sum of the discriminatory scores of all the features in the data set, that is $M_2=\sum_{i=1}^{m} F(i)$. We assume the condition that $M_2>0$ as $\sum_{i=1}^{m} F(i)>0$.

In the proposed algorithm we perform a stratified $k$ fold cross validation and split the data set into ten training and test data sets. For each training data set we extract the feature sub set using standard PSO and ADT classifier. We then compute the feature discrimination score using the formula~\ref{chap4:eqn1}. We compare the new feature score with the previous scores and the algorithm increases the particle population size in the standard PSO by a factor of one till a local maximum is found. To obtain the local maximum point we first obtain the first and second derivative of number of particles in the iteration $i$ (say $y$) and the feature discriminatory score $V_{i}$ (say $x$). The local maxima is computed as given in Equation~\ref{eqnder}. The loop is terminated when the conditions in Equation \ref{eqnder} are met.

\begin{eqnarray}
\label{eqnder}
\frac{dy}{dx}_{i-2} > \frac{dy}{dx}_{i-1} \&\& \nonumber \\
 \frac{dy}{dx}_{i-1} > \frac{dy}{dy}_{i} \&\& \frac{d^2 y}{dx^2}  < 0
\end{eqnarray}

\begin{algorithm}
\caption{New Feature Selection (NFS) Algorithm}\label{chap4:NFS1}
\begin{algorithmic}
\STATE {\bf Input:} Data sets for the purpose of decision making $S(n,m)$ where $n$ and $m$ are number of records and attributes respectively. 
\STATE {\bf Output:} Mean and standard deviation of the number of features selected and the classification accuracy $FM$, $FS$, $AM$, $AS$
   
   \begin{IEEEenumerate}
        \item Identify and collect all records in a data set $S$.
        \item Split the data set in to training and testing sets using a stratified  $k$ fold cross validation procedure. Denote each training and testing data set by $T_{k}$ and $R_{k}$ respectively.
   \item For each $k$
      \begin{IEEEenumerate}\label{NMproc}
      \item initialize the number of particles in PSO as $N$=$5$, the fitness value as $V_{0}=0$ and $i=0$.
			
       \begin{IEEEenumerate}\label{fsloop}
			
      \item Extract the optimal feature subset from $T_{k}$ using a wrapper based approach with PSO search for identifying feature subsets using number of particles as $N$ and alternating decision tree for its evaluation.\label{step1}
      \item Generate new training and testing data sets using the features identified in Step~\ref{step1} from $T_{k}$ and $R_{k}$ respectively. Designate these sets by $P_{i}$ and $Q_{i}$ respectively.\label{step2}
      \item Build the ADT using the training data set $P_{i}$ and obtain the test accuracy using testing data set $Q_{i}$. Designate this accuracy by $B_{i}$.\label{buildadt}
      \item Compute the feature scores of data sets $P_{i}$ and $T_{k}$ respectively. Designate the scores by $M_{1}$ and $M_{2}$ respectively.
      \item Compute the fitness value as $V_{i}=0.5*A+0.5*(M_{1}/M_{2}).$ \label{endstep}
      \item Update $N$ as $N+1$ and $i$ as $i+1$.
      \item check for local maxima using Equation \ref{eqnder} and exit the loop if conditions are met.
    \end{IEEEenumerate}
      \item Obtain the feature subset using the particles obtained in set above in PSO feature selection using ADT.
			\item Obtain the test accuracy $A_k$ by building an ADT using the features obtained from above and designate the accuracy of fold $k$ and $F_{k}=\#D_{i}$. \label{endkfold}
     \end{IEEEenumerate}

      \item  Repeat the Steps (3)-(i) to Step (3)-\ref{endkfold} for each fold.
        \item Designate $FM$, $FS$, $AM$, $AS$ as the mean and standard deviation of $F_{k}$ and $A_{k}$ respectively.
         \item RETURN $FM$, $FS$, $AM$, $AS$.
         \item END.
   \end{IEEEenumerate}
\end{algorithmic}
\end{algorithm}

\subsection{Validation}

To evaluate the performance of the proposed Tunable Particle Swarm Size Optimization Algorithm, we first obtain the feature subset corresponding to the number of particles found using the above procedure. We then train an ADT using the features identified in the previous step and then compute test accuracies on the test data set of the corresponding fold. The procedure is repeated for all the ten folds and the average accuracy is computed. The above procedure is given as Algorithm~\ref{chap4:NFS1}. 

\section{Experiments and Results}
\label{expresults}

We have conducted experiments on bench mark data sets obtained from University of California Irvin (UCI) data repository~\cite{Frank2010} StatLog project, Keel~\cite{Alcala-Fdez2010} and Bangor data repositories (https://www.bangor.ac.uk/). The performance of the proposed algorithm TPSO is compared with standard PSO and GA with alternating decision tree classifier. The description of the data sets are given in the Table~\ref{chap4:tab1}. 

\begin{table*}
\begin{center}
\caption{Data sets}
{\begin{tabular}{|l|l|l|c|c|}\hline 
Name          		& Source        & Number of  		&  Number of     & Number of 	 \\ 
			&		& Features		& Classes	 & Records 	\\ \hline \hline

Australian (AUS)	& UCI Statlog	&14			&2		 	&690\\ \hline
Contractions (CON)	& Bangor	&27			&2			&98\\ \hline
German (GER) 		&Keel		&20			&2			&1000\\ \hline
Heart (HRT) 		&UCI Statlog	&13			&2			&270\\ \hline
Ionosphere (ION) 	&UCI		&33			&2			&351\\ \hline
Laryngeal1 (LAR) 	& Bangor	&16			&2			&213\\ \hline
Respiratory 		&		&			&			&\\
distress syndrome (RDS) &Bangor		&17			&2			&85\\ \hline
Sonar (SNR)		& UCI		&60			&2			&208\\ \hline
WDBC (WBD) 		&Keel		&30			&2			&569\\ \hline
Weaning (WEA)		&Bangor		&17			&2			&302\\ \hline
\end{tabular}}
\label{chap4:tab1}
\end{center}
\end{table*}

In the present methodology we employ a stratified $k$-fold cross validation ($k=10$) procedure. The folds are selected so that the mean response value is approximately equal in all the folds. In case of a dichotomous classification, this means that each fold has roughly the same proportions of the two types of class labels. Table~\ref{parameters} provides details of the default necessary parameters of GA and PSO in the current experimental study.

\begin{table*}
\begin{center}
\caption{Parameter Setup}
{\begin{tabular}{|l||c|}\hline 
Name          		& Parameters       	 	 \\ \hline \hline

GA			& Crossover=1.0 \\
			& mutation probability=0.001		\\ \hline
PSO			& initial number of particles Z= 50 \\
			& iterations G=100 \\
			& cognitive factor $c_1=2$ \\
			& social factor $c_2=2$		\\ \hline
\end{tabular}}
\label{parameters}
\end{center}
\end{table*}

\subsection{Computational Complexity and Scalability}
\label{chap8:sec4}

The computational complexity is a measure of the performance of the algorithm. For each data set having $n$ attributes and $m$ records, we select only those subset of records $m_{1} \le m$, in which missing values are present. The distances  are computed for all attributes $n$ excluding the decision attribute. So, the time complexity for computing the distance would be $O(m_{1}*(n-1))$. The time complexity for selecting the nearest records is of order $O(m_{1})$. For computing the frequency of occurrences for nominal attributes and average for numeric attributes the time taken would be of the order $O(m_{1})$.  In case of the proposed method, let $O(p)$ be the time complexity of wrapper based feature sub set identification using standard PSO and ADT. For $F$ folds the complexity would be $F*O(p)$. For $I$ changes in the swarm size the time complexity of feature selection step would be $F*O(p)*I$. Therefore, for a given data set with $k$-fold cross validation having $n$ attributes and $m$ records, the time complexity of TPSO would be $k*(O(m_{1}*(n-1)*m)+2*O(m_{1})+F*O(p)*I)$ which is asymptotically linear.

\begin{figure}
\begin{center}
\includegraphics[width=3.0in]{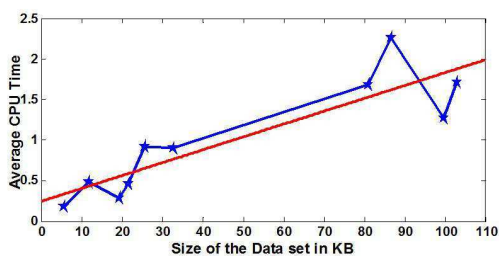}
\caption{Computational complexity of the TPSO algorithm}
\label{TCNFS}
\end{center}
\end{figure}

A plot between the varying sizes of the data sets and the time taken for processing by the proposed algorithm (TPSO) is shown in Fig.~\ref{TCNFS}. Also, we employed a linear regression on our results and obtained the relation between the time taken (T) and the data size (D) as $T=0.0159D+0.2464$, $\alpha=0.05$, $p<0.05$, $r^2=0.78$.

The presence of the linear trend between the time taken and the varying database sizes ensures the numerical scalability of the performance of TPSO in terms of asymptotic linearity.

\subsection{Performance Comparison on Benchmark Data sets}

Firstly, we compared the accuracy of the proposed TPSO method with accuracies of ADT classifier without employing any feature selection. The results are tabulated in Table~\ref{chap4:tab41}.

\begin{table*} 
\caption{Comparison of accuracies obtained by ADT without feature selection and TPSO}
\begin{center}
\begin{tabular}{|l|c|c|}\hline 
Dataset         & TPSO	   		& ADT  		 	 \\ \hline \hline
                   	  
AUS		& 87.54 $\pm$	3.82 	& 84.35 $\pm$ 3.33\\	\hline
CON		& 87.89	$\pm$ 10.31 	& 84.78 $\pm$ 12.81\\	\hline
GER		& 74.4 $\pm$ 3.75	& 72.8 $\pm$ 4.59		\\	\hline
HRT		& 83.33 $\pm$ 7.46	& 78.89 $\pm$ 10.48 \\	\hline
ION		& 94.86 $\pm$ 3.24	& 90.02 $\pm$ 4.92	\\	\hline
LAR		& 86.86 $\pm$ 6.91	& 82.23 $\pm$ 9.37\\	\hline
RDS		& 90.56 $\pm$ 11.37	& 90.56 $\pm$ 11.37	\\	\hline
SNR		& 86.95	$\pm$ 8.00	& 83.05 $\pm$ 9.23		\\	\hline
WBD		& 97.19 $\pm$ 3.12	& 94.38 $\pm$ 3.86	\\	\hline
WEA		& 87.42 $\pm$ 4.38	& 81.44 $\pm$ 6.35	\\	\hline
\end{tabular}
\label{chap4:tab41}
\end{center}
\end{table*}

Later, we considered GA and standard PSO algorithms for feature subset selection and ADT classifier as wrapper for feature evaluation.
A comparison of the performances of TPSO with GA+ADT and Standard PSO+ADT methods on benchmark data sets is shown in Table~\ref{chap4:tab4}.

\begin{table*}
\caption{Comparison of accuracies obtained by TPSO, GA+ADT, Standard PSO+ADT}
\begin{center}
\begin{tabular}{|l|c|c|c|}\hline 
Dataset         & TPSO 	   		& GA+ADT  		& Standard PSO+ADT	 	 \\  \hline \hline
                   	  
AUS		& 87.54 $\pm$3.82 	& 85.51 $\pm$3.86	& 84.49 $\pm$ 2.65	\\	 \hline
CON		& 87.89	$\pm$ 10.31 	& 77.78 $\pm$ 14.56	& 80.78 $\pm$ 9.99	 \\	\hline
GER		& 74.4 $\pm$ 3.75	& 69.8 $\pm$ 4.9	& 70.4 $\pm$ 5.64	 \\	\hline
HRT		& 83.33 $\pm$ 7.46	& 83.7	$\pm$ 7.24 	& 76.67 $\pm$ 9.08	 \\	\hline
ION		& 94.86 $\pm$ 3.24	& 89.75 $\pm$ 6.74	& 92.01 $\pm$ 3.77	 \\	\hline
LAR		& 86.86 $\pm$ 6.91	& 79.81 $\pm$ 7.62	& 79.83 $\pm$ 9.19	 \\	\hline
RDS		& 90.56 $\pm$ 11.37	& 90.56 $\pm$ 8.23	& 89.31 $\pm$ 10.9	 \\	 \hline
SNR		& 86.95	$\pm$ 8.00	& 81.19 $\pm$ 11.33	& 81.12 $\pm$ 10.10	\\	 \hline
WBD		& 97.19 $\pm$ 3.12	& 95.26 $\pm$ 3.7	& 95.25 $\pm$ 3.42	\\	 \hline
WEA		& 87.42 $\pm$ 4.38	& 84.78 $\pm$ 3.81	& 82.76 $\pm$ 5.88	 \\	\hline
\end{tabular}
\label{chap4:tab4}
\end{center}
\end{table*}

The mean and the standard deviation of the number of features selected for $k$ folds of the cross validation procedure is shown Table~\ref{chap4:tab5}.

\begin{table*}
\caption{Comparison of number of features selected using TPSO with GA+ADT and Standard PSO+ADT}
\begin{center}
\begin{tabular}{|l||c||c||c||c|}
\hline
Dataset	& Original	&TPSO			& GA+ADT		& Standard PSO+ADT \\ \hline \hline
AUS	& 14		& 3.8	$\pm$	2.20	& 5.3	$\pm$	1.77	& 2.8	$\pm$	 2.78	\\	 \hline
CON	& 27		& 13.2	$\pm$	5.12	& 9.5	$\pm$	2.46	& 11.5	 $\pm$	4.28	\\	 \hline
GER	& 20		& 10	$\pm$	2.31	& 7.8	$\pm$	2.04	& 8.5	$\pm$	 2.68	\\	 \hline
HRT	& 13		& 9.3	$\pm$	1.49	& 4.3	$\pm$	2.11	& 9.3	$\pm$	 1.16	\\	 \hline
ION	& 34		& 13.8	$\pm$	4.57	& 12.6	$\pm$	2.63	& 11.2	 $\pm$	3.88	\\	 \hline
LAR	& 16		& 4.9	$\pm$	2.18	& 5.4	$\pm$	2.12	& 6.8	$\pm$	 2.1	\\	 \hline
RDS	& 17		& 10.0	$\pm$	2.26	& 6.4	$\pm$	1.35	& 9.3	 $\pm$	1.49	\\	 \hline
SNR	& 60		& 33.8	$\pm$	10.21	& 28.6	$\pm$	4.65	& 35.0	 $\pm$	9.93	\\	 \hline
WBD	& 30		& 13.4	$\pm$	3.06	& 9.4	$\pm$	2.84	& 13.0	 $\pm$	2.36	\\	 \hline
WEA	& 17		& 9.6	$\pm$	2.17	& 8.3	$\pm$	1.25	& 9.7	$\pm$	 1.89	\\	 \hline
\end{tabular}
\label{chap4:tab5}
\end{center}
\end{table*}

From Table~\ref{chap4:tab5} it can be observed that the TPSO methodology has rendered higher accuracies using less than $50\%$ of the original set of attributes.

To substantiate the improvement in classification accuracy using TPSO methodology a statistical test based on Wilcoxon method is employed and the results are presented in Table~\ref{chap4:tab8}.

\begin{table*}
\caption {Wilcoxon sign rank statistics for matched pairs comparing TPSO with other feature selection methods}
\begin{center}
\begin{tabular}{|l||c||c||c||c|}
\hline
Method			& Rank Sums 	& Test		&  Critical	& p-value \\
			& (+, -)	& Statistics	& Value		&		\\ \hline \hline
GA+ADT			& 54.0,1.0	& 1.0		& 9.0		& 0.003\\ \hline
PSO+ADT			& 55.0,0.0	& 0.0		& 9.0		& 0.001\\ \hline
\end{tabular}
\label{chap4:tab8}
\end{center}
\end{table*}

From the Table~\ref{chap4:tab8} we infer that TPSO is superior to the standard PSO feature selection method with positive rank sum of $55$, $p<0.001$ and $\alpha=0.05$ significance. The TPSO method indicating a remarkable performance when compared with GA feature selection with a positive rank sum of $54$, $p<0.003$ and $\alpha=0.05$

\section{Conclusions and Discussion}
\label{concl}

In this paper, we have discussed the issues related to high dimensionality of the data sets and feature selection as a solution to the curse of dimensionality. The feature selection methods such as filter and wrapper have been discussed. 
Particle Swarm Optimization (PSO) is a population based optimization technique, which has been proved to get optimal feature subset provided the necessary input parameters are properly tuned. Particle swarm size is the critical parameter in standard PSO. To address this issue, we proposed a novel {\em tunable swarm size configuration} approach to find the population size of the particles based on the data sets in real time. 
The proposed algorithm is named as {\em Tunable Particle Swarm Size Optimization Algorithm (TPSO)}. A new fitness function has been developed which integrates the accuracy of the classifier with the modified F-Score. Empirically, we compare the performance of our new algorithm with other state-of-the-art classifiers on bench marking data sets obtained from UCI, Keel and Bangor data repositories. Wilcoxon statistical test confirmed the fact that the proposed algorithm has improved the classification accuracies in comparison to other methods.



\bibliographystyle{IEEEtran}

\begin{thebibliography}{10}
\providecommand{\url}[1]{#1}
\csname url@samestyle\endcsname
\providecommand{\newblock}{\relax}
\providecommand{\bibinfo}[2]{#2}
\providecommand{\BIBentrySTDinterwordspacing}{\spaceskip=0pt\relax}
\providecommand{\BIBentryALTinterwordstretchfactor}{4}
\providecommand{\BIBentryALTinterwordspacing}{\spaceskip=\fontdimen2\font plus
\BIBentryALTinterwordstretchfactor\fontdimen3\font minus
  \fontdimen4\font\relax}
\providecommand{\BIBforeignlanguage}[2]{{%
\expandafter\ifx\csname l@#1\endcsname\relax
\typeout{** WARNING: IEEEtran.bst: No hyphenation pattern has been}%
\typeout{** loaded for the language `#1'. Using the pattern for}%
\typeout{** the default language instead.}%
\else
\language=\csname l@#1\endcsname
\fi
#2}}
\providecommand{\BIBdecl}{\relax}
\BIBdecl

\bibitem{bioucas2013hyperspectral}
J.~M. Bioucas-Dias, A.~Plaza, G.~Camps-Valls, P.~Scheunders, N.~Nasrabadi, and
  J.~Chanussot, ``Hyperspectral remote sensing data analysis and future
  challenges,'' \emph{IEEE Geoscience and remote sensing magazine}, vol.~1,
  no.~2, pp. 6--36, 2013.

\bibitem{hua2009performance}
J.~Hua, W.~D. Tembe, and E.~R. Dougherty, ``Performance of feature-selection
  methods in the classification of high-dimension data,'' \emph{Pattern
  Recognition}, vol.~42, no.~3, pp. 409--424, 2009.

\bibitem{liu2002}
H.~Liu, J.~Li, and L.~Wong, ``A comparative study on feature selection and
  classification methods using gene expression profiles and proteomic
  patterns,'' \emph{GENOME INFORMATICS SERIES}, pp. 51--60, 2002.

\bibitem{Guyon2003}
I.~Guyon and A.~Elisseeff, ``An introduction to variable and feature
  selection,'' \emph{Journal of Machine Learning Research}, vol.~3, pp.
  1157�--1182, 2003.

\bibitem{Guyon2002}
I.~Guyon, J.~Weston, S.~Barnhill, and V.~Vapnik, ``Gene selection for cancer
  classification using support vector machines,'' \emph{Machine Learning},
  vol.~46, pp. 389--422, 2002.

\bibitem{Liu2005}
H.~Liu, E.~R. Dougherty, J.~G. Dy, K.~Torkkola, E.~Tuv, H.~Peng, C.~Ding,
  F.~Long, M.~Berens, L.~Parsons, Z.~Zhao, L.~Yu, and G.~Forman, ``Evolving
  feature selection,'' \emph{IEEE Intelligent Systems}, vol.~20, pp. 64--76,
  2005.

\bibitem{yang2012particle}
H.~Yang, Q.~Du, and G.~Chen, ``Particle swarm optimization-based hyperspectral
  dimensionality reduction for urban land cover classification,'' \emph{IEEE
  Journal of Selected Topics in Applied Earth Observations and Remote Sensing},
  vol.~5, no.~2, pp. 544--554, 2012.

\bibitem{Inza2004}
I.~n. Inza, P.~Larra\~{n}aga, R.~Blanco, and A.~J. Cerrolaza, ``Filter versus
  wrapper gene selection approaches in dna microarray domains.''
  \emph{Artificial intelligence in medicine}, vol. 31(2), pp. 91--103, 2004.

\bibitem{Forman2003}
F.~George, ``An extensive empirical study of feature selection metrics for text
  classification,'' \emph{Journal of Machine Learning Research}, vol.~3, pp.
  1289--1305, 2003.

\bibitem{Swets1995}
D.~Swets and J.~Weng, ``Efficient content-based image retrieval using automatic
  feature selection,'' in \emph{IEEE International Symposium On Computer
  Vision}, 1995, pp. 85--90.

\bibitem{Lee2000}
W.~Lee, S.~J. Stolfo, and K.~W. Mok, ``Adaptive intrusion detection: A data
  mining approach,'' \emph{AI Review}, vol. 14(6), pp. 533--567, 2000.

\bibitem{Saeys2007}
Y.~Saeys, I.~Inza, and P.~LarrANNaga, ``A review of feature selection
  techniques in bioinformatics,'' \emph{Bioinformatics}, vol. 23(19), pp.
  2507--2517, 2007.

\bibitem{chandrashekar2014survey}
G.~Chandrashekar and F.~Sahin, ``A survey on feature selection methods,''
  \emph{Computers \& Electrical Engineering}, vol.~40, no.~1, pp. 16--28, 2014.

\bibitem{Tao2004}
L.~Tao, Z.~Chengliang, and O.~Mitsunori, ``A comparative study of feature
  selection and multiclass classification methods for tissue classification
  based on gene expression,'' \emph{Bioinformatics}, vol. 20(15), pp.
  2429--2437, 2004.

\bibitem{sun2005}
Y.~Sun, C.~F. Babbs, and E.~J. Delp, ``A comparison of feature selection
  methods for the detection of breast cancers in mammograms: Adaptive
  sequential floating search vs. genetic algorithm,'' in \emph{Engineering in
  Medicine and Biology Society, 2005. IEEE-EMBS 2005. 27th Annual International
  Conference of the}, 2005, pp. 6532--6535.

\bibitem{Shuangge2006}
S.~Ma, ``Empirical study of supervised gene screening,'' \emph{BMC
  Bioinformatics}, vol.~7, pp. 537+, 2006.

\bibitem{Murie2009}
M.~Carl, W.~Owen, L.~Anna, and N.~Robert, ``Comparison of small n statistical
  tests of differential expression applied to microarrays,'' \emph{BMC
  Bioinformatics}, vol. 10(1), p.~45, 2009.

\bibitem{raymer2000dimensionality}
M.~L. Raymer, W.~F. Punch, E.~D. Goodman, L.~A. Kuhn, and A.~K. Jain,
  ``Dimensionality reduction using genetic algorithms,'' \emph{IEEE
  transactions on evolutionary computation}, vol.~4, no.~2, pp. 164--171, 2000.

\bibitem{xue2012new}
B.~Xue, M.~Zhang, and W.~N. Browne, ``New fitness functions in binary particle
  swarm optimisation for feature selection,'' in \emph{Evolutionary Computation
  (CEC), 2012 IEEE Congress on}.\hskip 1em plus 0.5em minus 0.4em\relax IEEE,
  2012, pp. 1--8.

\bibitem{shi1998}
Y.~Shi and R.~C. Eberhart, ``{Parameter Selection in Particle Swarm
  Optimization},'' in \emph{EP '98: Proceedings of the 7th International
  Conference on Evolutionary Programming VII}.\hskip 1em plus 0.5em minus
  0.4em\relax London, UK: Springer-Verlag, 1998, pp. 591--600.

\bibitem{angeline1998}
P.~Angeline, ``Evolutionary optimization versus particle swarm optimization:
  Philosophy and performance differences,'' in \emph{Evolutionary programming
  VII}.\hskip 1em plus 0.5em minus 0.4em\relax Springer, 1998, pp. 601--610.

\bibitem{xue2016survey}
B.~Xue, M.~Zhang, W.~N. Browne, and X.~Yao, ``A survey on evolutionary
  computation approaches to feature selection,'' \emph{IEEE Transactions on
  Evolutionary Computation}, vol.~20, no.~4, pp. 606--626, 2016.

\bibitem{vandenBergh2001}
F.~van~den Bergh and A.~P. Engelbrecht, ``Effects of swarm size on cooperative
  particle swarm optimisers,'' in \emph{Proceedings of the Genetic and
  Evolutionary Computation Conference (GECCO)}, 2001, pp. 892--�899.

\bibitem{Quinlan1993}
J.~R. Quinlan, \emph{C4.5: programs for machine learning}.\hskip 1em plus 0.5em
  minus 0.4em\relax San Francisco, CA, USA: Morgan Kaufmann Publishers Inc.,
  1993.

\bibitem{Efron2004}
B.~Efron, T.~Hastie, I.~Johnstone, and R.~Tibshirani, ``Least angle
  regression,'' \emph{Annals of Statistics}, vol. 32(2), pp. 407--499, 2004.

\bibitem{Hastie2003}
T.~Hastie, R.~Tibshirani, and J.~Friedman, \emph{The Elements of Statistical
  Learning: Data Mining, Inference, and Prediction}.\hskip 1em plus 0.5em minus
  0.4em\relax Springer, 2003.

\bibitem{jin2006machine}
X.~Jin, A.~Xu, R.~Bie, and P.~Guo, ``Machine learning techniques and chi-square
  feature selection for cancer classification using sage gene expression
  profiles,'' in \emph{International Workshop on Data Mining for Biomedical
  Applications}.\hskip 1em plus 0.5em minus 0.4em\relax Springer, 2006, pp.
  106--115.

\bibitem{Wang2007}
S.~Wang, C.-L. Liu, and L.~Zheng, ``Feature selection by combining fisher
  criterion and principal feature analysis,'' in \emph{Machine Learning and
  Cybernetics, 2007 International Conference on}, vol.~2, 2007, pp. 1149--1154.

\bibitem{Liu2004}
Y.~Liu, ``A comparative study on feature selection methods for drug
  discovery,'' \emph{Journal of Chemical Information and Computer Sciences},
  vol. 44(5), pp. 1823--1828, 2004.

\bibitem{peng2005}
H.~Peng, F.~Long, and C.~Ding, ``Feature selection based on mutual information
  criteria of max-dependency, max-relevance, and min-redundancy,'' \emph{IEEE
  Transactions on Pattern Analysis and Machine Intelligence}, vol. 27(8), pp.
  1226--1238, 2005.

\bibitem{Tan2008}
Q.~Tan, M.~Thomassen, K.~Jochumsen, J.~Zhao, K.~Christensen, and T.~Kruse,
  ``Evolutionary algorithm for feature subset selection in predicting tumor
  outcomes using microarray data,'' in \emph{Bioinformatics Research and
  Applications}, ser. Lecture Notes in Computer Science, I.~Mandoiu,
  R.~Sunderraman, and A.~Zelikovsky, Eds.\hskip 1em plus 0.5em minus
  0.4em\relax Springer Berlin / Heidelberg, 2008, vol. 4983, pp. 426--433.

\bibitem{winkler2012analysis}
S.~Winkler, M.~Affenzeller, G.~Kronberger, M.~Kommenda, S.~Wagner, W.~Jacak,
  and H.~Stekel, ``Analysis of selected evolutionary algorithms in feature
  selection and parameter optimization for data based tumor marker modeling,''
  \emph{Computer Aided Systems Theory--EUROCAST 2011}, pp. 335--342, 2012.

\bibitem{Malin2008}
M.~B. {\AA}berg, L.~L{\"o}ken, and J.~Wessberg, ``An evolutionary approach to
  multivariate feature selection for fmri pattern analysis,'' in
  \emph{BIOSIGNALS (2)}, 2008, pp. 302--307.

\bibitem{tran2016investigation}
B.~Tran, B.~Xue, M.~Zhang, and S.~Nguyen, ``Investigation on particle swarm
  optimisation for feature selection on high-dimensional data: Local search and
  selection bias,'' \emph{Connection Science}, vol.~28, no.~3, pp. 270--294,
  2016.

\bibitem{lynn2017population}
N.~Lynn, M.~Z. Ali, and P.~N. Suganthan, ``Population topologies for particle
  swarm optimization and differential evolution,'' \emph{Swarm and Evolutionary
  Computation}, 2017.

\bibitem{freund1999alternating}
Y.~Freund and L.~Mason, ``The alternating decision tree learning algorithm,''
  in \emph{icml}, vol.~99, 1999, pp. 124--133.

\bibitem{naresh2013}
M.~N. Kumar, ``Alternating decision trees for early diagnosis of dengue
  fever,'' \emph{arXiv preprint arXiv:1305.7331}, 2013,.

\bibitem{liu2011improved}
Y.~Liu, G.~Wang, H.~Chen, H.~Dong, X.~Zhu, and S.~Wang, ``An improved particle
  swarm optimization for feature selection,'' \emph{Journal of Bionic
  Engineering}, vol.~8, no.~2, pp. 191--200, 2011.

\bibitem{Frank2010}
\BIBentryALTinterwordspacing
A.~Frank and A.~Asuncion, ``{UCI} machine learning repository,'' 2010.
  [Online]. Available: \url{http://archive.ics.uci.edu/ml}
\BIBentrySTDinterwordspacing

\bibitem{Alcala-Fdez2010}
A.~Alcal�-Fdez, A.~Fernandez, Luengo, J.~Derrac, S.~G. J., L.~S�nchez, and
  F.~Herrera, ``Keel data-mining software tool: Data set repository,
  integration of algorithms and experimental analysis framework,''
  \emph{Journal of Multiple-Valued Logic and Soft Computing}, 2010.

\end{thebibliography}
%

\end{document}